\documentclass[11pt]{article}
\usepackage{coling2018}
\usepackage{times}
\usepackage{url}
\usepackage{latexsym}
\usepackage{latexsym}
\usepackage{url}
\usepackage{amsmath}
\usepackage{booktabs}
\usepackage[dvipsnames]{xcolor}
\usepackage{graphicx}
\usepackage{tikz}
\usepackage{pgfplots}
\newcommand{\topline}{\toprule}
\newcommand{\midline}{\midrule}
\newcommand{\bottomline}{\bottomrule}
\newcommand{\blue}[1]{\textcolor{black}{#1}}

\newcommand{\added}[1]{\textcolor{black}{#1}}
\newcommand{\addedd}[1]{\textcolor{black}{#1}}
\newcommand{\addet}[1]{\textcolor{black}{#1}}

\newcommand{\citet}[1]{\newcite{#1}}

\title{Word-Level Loss Extensions for Neural Temporal Relation Classification}
\author{Artuur Leeuwenberg \and Marie-Francine Moens\\Department of Computer Science\\KU Leuven, Belgium\\\textit{\{tuur.leeuwenberg, sien.moens\}@cs.kuleuven.be}
}

\date{}
 
\begin{document}

\maketitle

\blfootnote{\hspace{-0.65cm}  
     This work is licensed under a Creative Commons 
     Attribution 4.0 International License.
     License details:
     \url{http://creativecommons.org/licenses/by/4.0/}}
     
\begin{abstract}
Unsupervised pre-trained word embeddings are used effectively for many tasks in natural language processing to leverage unlabeled textual data. Often these embeddings are either used as initializations or as fixed word representations for task-specific classification models.
In this work, we extend our classification model's task loss with an unsupervised auxiliary loss on the word-embedding level of the model. This is to ensure that the learned word representations contain both task-specific features, learned from the supervised loss component, and more general features learned from the unsupervised loss component. We evaluate our approach on the task of temporal relation extraction, in particular, narrative containment relation extraction from clinical records, and show that continued training of the embeddings on the unsupervised objective together with the task objective gives better task-specific embeddings, and results in an improvement over the state of the art on the THYME dataset, using only a general-domain part-of-speech tagger as linguistic resource.
\end{abstract}

\noindent

\section{Introduction}

Word representations in the form of continuous vectors are often pre-trained on large amounts of raw text to learn general word features, using unsupervised objectives. These representations are then used in supervised models for various classification tasks. However, such tasks sometimes require very specific features that may not have been captured by the unsupervised objective. In other domains such as computer vision, representations are often learned jointly from multiple resources for classification. In this work, we explore the possibility to exploit learning signals from both settings to construct better task-oriented word representations, and obtain a better relation classification model.

\begin{figure}[h!]
\centering
\resizebox{.6\textwidth}{!}{
\includegraphics[]{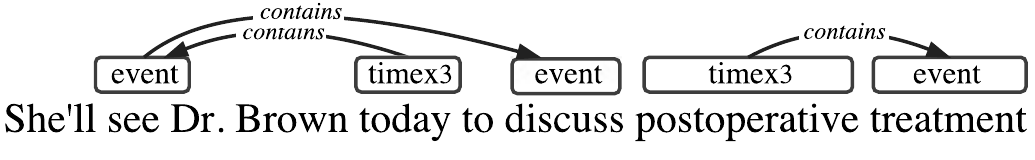}
}
\caption{\label{fig:example} A sentence annotated with events, temporal expressions (Timex3), and containment relations.}
\end{figure}

The main task in this work is the extraction of narrative containment relations (CR) from English clinical texts, as annotation of clinical data is costly and it is therefor crucial to fully exploit both the labeled as well as the unlabeled data that is available. The aim of CR extraction is to find if, given events A and B, event A is temporally contained in event B (i.e. if event A happens within the time span of event B). An example of such relation is given in Fig. \ref{fig:example}, \added{where the model should predict all containment edges given the entities (events and temporal expressions), by classifying each pair of entities as containment or no containment}.
Temporal relation classification in clinical text is a very important task in the secondary use of clinical data from electronic health records. The patient time-line is crucial for making a good patient prognosis and clinical decision support \cite{Onisko2015}.
This task has already been addressed in three iterations of the Clinical TempEval Shared Task \cite{bethard2016semeval}. Still there is a gap of more than 0.20 in F-measure between the state-of-the-art CR extraction systems and the inter-adjunctator agreement (indicating an upper bound for performance). This shows that this task is very challenging.

\addedd{Following the current trend in NLP, the recent state-of-the-art models for extraction of CR are neural network models. These models all use pre-trained word embeddings as word representations \cite{tourille2017bilstm,dligach2017neural,lin2017representations}.}

Pre-training of the embeddings is done with an auxiliary task (a task where one is not interested in the final predictions, but in the trained model components), like the skip-gram task \cite{mikolov2013efficient}.
When used for classification tasks in NLP, these pre-trained word representations are often either used as fixed inputs for the classification model, or as initialization for the word representations of the classification model (sometimes called fine-tuned embeddings).

A problem with pre-trained representations in classification models is that solving the main task often requires different information than the auxiliary task. Training word representations only on the auxiliary objective can result in loss of crucial information for the task, and afterwards fine-tuning on the task loss does not influence words that are not in the task's training data, losing generalization to those words. 

In the current work, we propose a neural relation classification (RC) model that learns its word representations jointly on the main task (supervised, on labeled data) and on the auxiliary task (unsupervised, on unlabeled data) in a multi-task setting to overcome this problem, and ensure that the embeddings contain valuable information for our main task, while still leveraging the unlabeled data for more general feature learning.
As auxiliary task we implement a skip-gram (SG) architecture, similar to \citet{mikolov2013efficient}.
Our proposed models use only unlabeled data and a general (news, out of domain) part-of-speech (POS) tagger as external resources, in contrast to the current state-of-the-art models, to ease extension to other languages for which specialized NLP tools for clinical texts might not be available.
The main contributions of this work are that it:
\begin{itemize}
\item Shows that training the word-level representations jointly on its main task and an auxiliary objective results in better \added{representations for} classification, compared to using pre-trained variants.
\item Shows that the method's increased performance and hyper-parameters are robust across different training set sizes, and that single-loss training settings act as lower bounds on performance.
\item Constitutes a new state of the art for temporal relation extraction on the THYME dataset even without dedicated clinical \mbox{preprocessing}.
\end{itemize}

\section{Related Work}

The model we present draws inspiration from prior research on (temporal) relation classification and neural multi-task learning.

\subsection{Clinical Temporal Relation Extraction}
Temporal relation extraction from clinical texts is a widely studied area in NLP and has been explored through various shared tasks, such as the i2b2 shared task on clinical temporal information \cite{sun2013evaluating}, and three iterations of Clinical TempEval \cite{bethard2015semeval,bethard2016semeval,bethard17semeval}. Until recently, most of the top performing systems employed manually constructed linguistic feature sets \cite{lin2015multilayered,lee2016uthealth,leeuwenberg2017structured}. In the last few years, there has been a shift towards using neural models, using LSTM \cite{tourille2017bilstm} and CNN models \cite{dligach2017neural,lin2017representations} inspired by the work on relation classification in other domains \cite{zeng2014relation,zhang2015relation,zhou2016attention,nguyen2015relation}. The top results in clinical temporal relation extraction are still achieved when enhancing the neural models with dedicated clinical NLP tools for preprocessing the clinical texts, often using the English cTAKES system \cite{savova2010mayo}, \added{which contains tools for clinical POS tagging, named entity recognition, and a dependency parser all trained on clinical data}. The main reason for using these dedicated clinical tools is that parsers trained on non-clinical texts perform significantly worse on clinical data \cite{jiang2015parsing}. Dedicated clinical NLP tools are not available for most languages though, and retraining NLP tools on clinical data is quite resource intensive, \added{because it requires extra annotation effort. Additionally, clinical data is often difficult to obtain or share publicly for patient privacy reasons}. Hence, we keep resource intensive preprocessing to a minimum and employ only a general news domain POS tagger \cite{toutanova2003feature}, providing important temporal relation extraction cues, such as tense shifts \cite{derczynski2016automatically}, and for which training data are available for many languages \cite{tagsetlrec2012}.




\subsection{Multi-task Learning}
Our proposed model training can be seen as multi-task learning (MTL), where the aim is to improve model generalization by leveraging the information from training signals of different related tasks \cite{caruana1998multitask}.
In earlier work, MTL has shown to be quite effective for different NLP tasks such as machine translation \cite{dong2015multi}, sentiment analysis \cite{peng2015named,yu2016learning}, sentence level name prediction \cite{cheng2015open}, semantic role labeling \cite{collobert2008unified}, and many more. For example, \citet{collobert2008unified} used an auxiliary unsupervised objective for semantic role labeling (SRL). They alternately trained embeddings in a language model and a SRL model. In contrast to their work, we learn both tasks truly jointly, and optimize a single semi-supervised objective. Typically in neural MTL, one or more layers of the network are shared among different models. \addedd{Two issues in MTL are (1) how to determine if the tasks are related enough to benefit from each other, and (2) what layers to share among the models.} \citet{baxter2000model} theoretically argue that tasks are related when they share an inductive bias. In our model, we expect that the skip-gram task \cite{mikolov2013efficient} can act as a reasonable word-level inductive bias for our task, as it has already shown its effectiveness in SRL \cite{collobert2008unified} and sentiment analysis \cite{peng2015named} in MTL, and for many NLP classification tasks when using them as pre-trained embeddings. \citet{hashimoto2016joint} showed that even when combining many tasks, considering the task hierarchy (simpler tasks lower in the network) allows them to benefit from each other. In most work on MTL the auxiliary tasks are supervised and specifically chosen for their relatedness to the main task \cite{ruder2017overview}, whereas in our model we chose the unsupervised auxiliary skip-gram task, and share weights of the word embedding layer. This results in a new joint relation classification objective that is semi-supervised on the word-level and provides better generalization for the final classification model. 


\section{The Model}
\begin{figure}[h!]
\centering
\resizebox{.5\textwidth}{!}{
\includegraphics[]{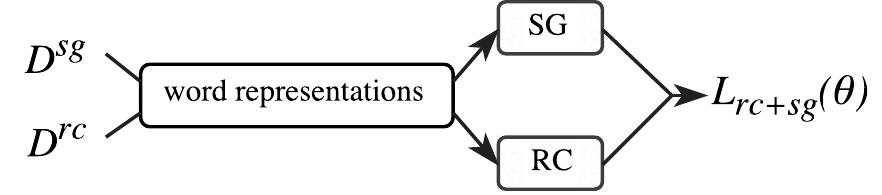}
}
\caption{\label{fig:high_level_model} High-level overview of our model training setting. $D^{rc}$ and $D^{rc}$ indicate the dataset for the relation classification and skip-gram components respectively, and $L_{rc+sg}(\theta)$ the model's combined loss.}
\end{figure}

Our model consist of two components: (1) a relation classification component (RC), and (2) a skip-gram component (SG). A high-level schematic overview of our model's training setting is shown in Fig. \ref{fig:high_level_model}.

\subsection{Relation Classification (RC)}
\begin{figure}[h!]
\centering
\resizebox{\textwidth}{!}{
\includegraphics[]{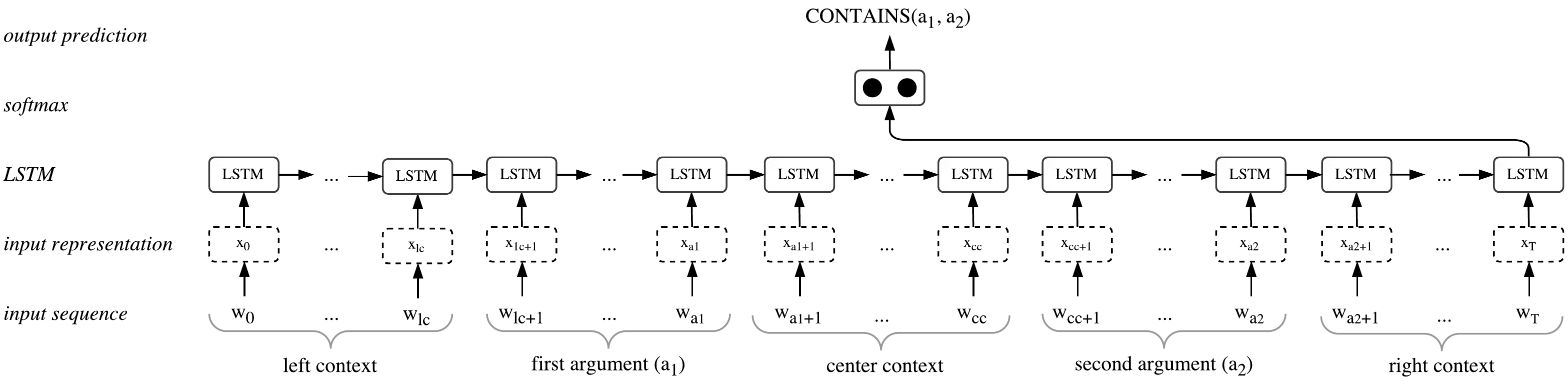}
}
\caption{\label{fig:lstm} Schematic representation of the relation classification (RC) model component. Arrows represent sets of fully connected weights. The dashed box indicates a word input as shown in Fig. \ref{fig:word_input}.}
\end{figure}

To classify relations we employ a long short-term memory (LSTM) \cite{hochreiter1997long} relation classification model (RC) \cite{zhang2015relation}. \addedd{We frame} the task as a sequence classification problem, taking as an input: the textual candidate relation description, i.e. the arguments of the candidate relation (the entity pair), and the context words surrounding the arguments, all read as a sequence from left to right. A schematic overview of the RC model component is shown in Fig. \ref{fig:lstm}. 

\begin{figure}
\centering
\resizebox{.35\textwidth}{!}{
\includegraphics[]{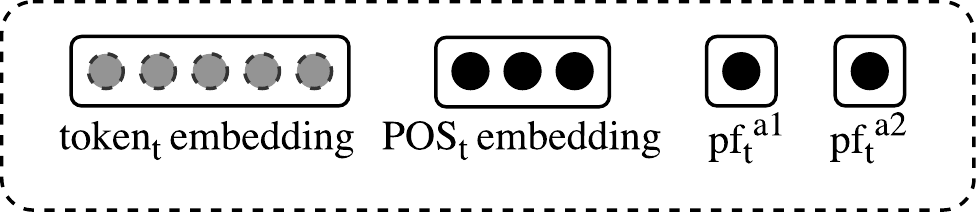}
}
\caption{\label{fig:word_input} Each word input $x_t$ of the RC model (at time step $t$) is a concatenation of a token embedding, a POS embedding, and two positional features (one for each argument).}
\end{figure}

Generation of candidate entity pairs is described later on in section \ref{sec:preproc}. The locations of the arguments \addedd{of each candidate relation} are indicated by two types of features taken from the literature: (1) position indicators, which are XML tags added to the original input sequence indicating the start and end of the arguments \cite{zhang2015relation}, and (2) by position features, indicating the \addedd{relative token-distance of each word to each argument} \cite{zeng2014relation}.
Each word's total input $x_t$ at time step $t\in\langle 0, 1, ..., T\rangle$ consists of the two argument locating features, $pf^{a1}_t$ and $pf^{a2}_t$, together with a word embedding $x^{token}_t \cdot W^{token}_{em}$, and a POS embedding $x^{pos}_t \cdot W^{pos}_{em}$, where $W^{token}_{em}$ and $W^{pos}_{em}$ are the embedding matrices for tokens and POS respectively, and $x^{token}_t$ and $x^{pos}_t$ their one-hot representations. A schematic overview of the concatenated input $x_t$ for each word to the LSTM unit is shown in \mbox{Fig. \ref{fig:word_input}}.


The predicted class probabilities $\hat{p}_{rc}(x)$ are given by a softmax classifier placed on top of the LSTM output $h$ at the last time step $T$ (in \mbox{Eq. \ref{eq:classifier}}).\footnote{We also experimented with bidirectional LSTMs \cite{zhang2015bidirectional} and adding attention \cite{zhou2016attention}. In our experiments, this did not result in significant improvements.}
\begin{align}
\label{eq:classifier}
\hat{p}_{rc}(x) &= \text{\textit{softmax}}(W_p h_T + b_p)
\end{align}
The RC model's loss function is cross-entropy loss, as shown in Eq. \ref{eq:cross_entropy}. \addedd{$D^{rc}$ indicates the supervised relation classification dataset}, and $\theta^{rc}$ is the collection of all trainable parameters of the model.
\begin{align}
\label{eq:cross_entropy}
L_{rc}(\theta^{rc}) = -\sum_{i=1}^{|D^{rc}|} y_i \log \hat{p}_{rc}(x_i)
\end{align}

\subsection{Context Prediction (SG)}As the unsupervised auxiliary task, we implemented a feed-forward neural network for a word context prediction task, known as the continuous skip-gram (SG), following \citet{mikolov2013efficient}. As input, the model takes a one-hot encoded input word $w_j$, which is projected to a word embedding, from which the probability distribution $y$ over its surrounding context words $w_{j-c}, ..., w_{j-1},w_{j+1}, ..., w_{j+c}$ is predicted, given a context window size $c$. The full model is given by Eq. \ref{eq:sg}.
\begin{align}
\label{eq:sg}
\hat{p}_{sg}(w_j) = softmax(W_{p_{sg}} (w_j\cdot W^{token}_{em}) + b_{p_{sg}})
\end{align}
Like the RC model, we use cross-entropy loss for our SG model, as shown in Eq. \ref{eq:cross_entropy_sg}. \addedd{$D^{sg}$ indicates the unsupervised dataset, consisting of words and their contexts.} $\theta^{sg}$ is the collection of all trainable parameters of our model.
\begin{align}
\label{eq:cross_entropy_sg}
L_{sg}(\theta^{sg}) = -\sum_{i=1}^{|D^{sg}|} y_i \log \hat{p}_{sg}(w_i)
\end{align}


\subsubsection{Separate Left \& Right Context (SGLR)}
\addedd{The skip-gram model is quite rough in its context description and does not take into account word order very well. However, for temporal relations we expect word order to be relevant. For this reason, we also experimented with a variation on the skip-gram model, separating the left and right context, following the intuition of \citet{ling2015two}.} The context separation is achieved by extending the context words by a `left' or `right' prefix depending on their location relative to the sampled word. 




\subsection{Combination (RC + SG)}
We train our proposed model on a combination of both loss functions, each with their own dataset $D^{rc}$, and $D^{sg}$ respectively. The combined loss, shown in Eq. \ref{eq:combined_loss}, is a weighted sum of their cross-entropy losses, where $\lambda_{sg}$ determines the importance of the SG loss.
\begin{align}
\label{eq:combined_loss}
L_{rc+sg}(\theta) = L_{rc}(\theta^{rc}) + \lambda_{sg} L_{sg}(\theta^{sg})
\end{align}

\begin{figure}[h!]

\resizebox{.55\textwidth}{!}{
\includegraphics[]{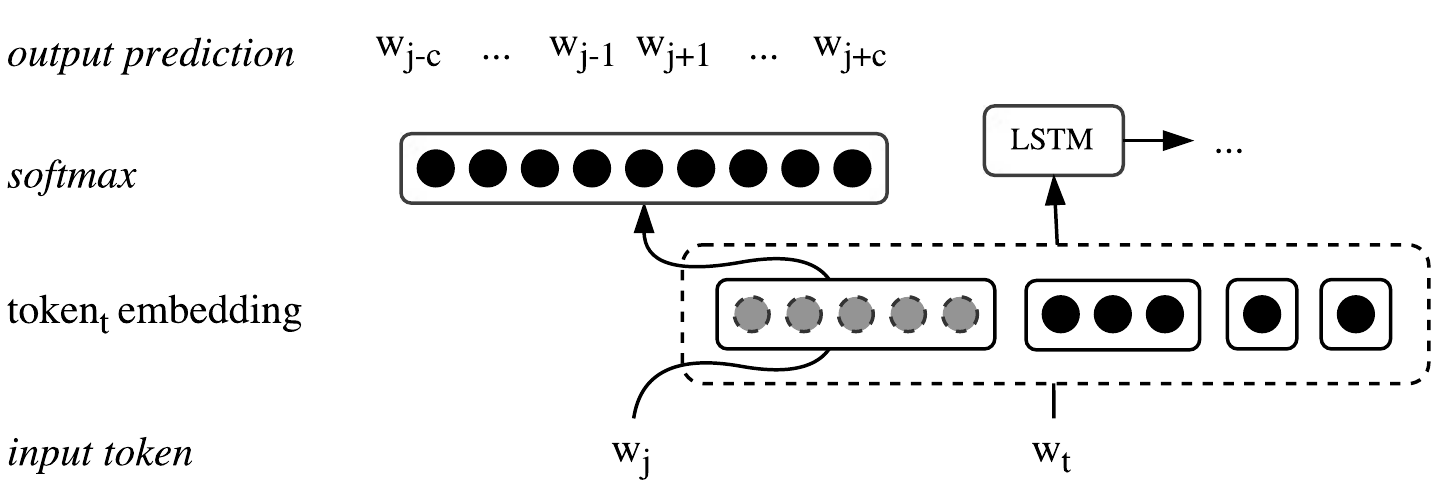}
}
\caption{\label{fig:combined_model} \addedd{Schematic representation of how the SG model component extends the RC model when using the combined loss on input word $w_j\in D^{sg}$, and word $w_t$ at time step $t$ from input sequence $x_i\in D^{rc}$. The gray layer indicates the shared word embedding parameters.} The dashed box represents the total word input $x_t$ for RC, as in Fig. \ref{fig:lstm} and \ref{fig:word_input}.}
\end{figure}

\added{A crucial part of our model is that although both models sample different types of inputs (\addet{the RC: sequences, the SG: single words}) from different datasets, and have different classification weights, the word embeddings are shared, i.e. $W^{token}_{em}$ ($\theta^{rc} \cap \theta^{sg} = W^{token}_{em}$), also illustrated in Fig. \ref{fig:high_level_model}. So only the word embeddings are directly influenced by both losses. All other weights (from RC or SG) are only influenced indirectly, through the word embedding weights, as both models are trained simultaneously.} \citet{sogaard2016deep} showed that, for NLP, sharing representations at the lower levels of the network is most effective: when lower level features are shared, there is room for the model to learn task specific abstractions in higher layers. For this reason we choose our model to share only the word embedding layer, as schematically illustrated in Fig. \ref{fig:combined_model}.


\subsection{Training}
We train all our models for at least 10 epochs, using Adam \cite{kingma2014adam} stochastic gradient
descent, with the default parameters from the original paper (lr=0.001), and a batch size of 1024 on a Titan X GPU. As stopping criteria we employ early stopping \cite{morgan1990generalization} with a patience of 20 epochs, based on F-measure on a small validation set of 3 documents (from Train). After each epoch on the main task, we shuffle the data and start the next one.
During training we employ a dropout of 0.5 on the input, and on the second last layer \cite{srivastava2014dropout}.

For our \addet{semi-supervised} setting, with the combined loss, we 
sample 1024 samples in our batch for each task from the corresponding  dataset, and do a single weight update on the combined loss.
\addet{A high-level schematic overview of our semi-supervised training setting is shown in Fig. \ref{fig:high_level_model}.}

\section{Experimental Setup}

\subsection{Datasets}

We conduct our experiments on the THYME corpus \cite{styler2014temporal}, a temporally annotated corpus of clinical notes in the colon cancer domain, also used in the Clinical TempEval Shared Tasks \cite{bethard2015semeval,bethard2016semeval}. We use the provided Train, Dev and Test split so we can directly compare to other approaches from the literature. The Train section consists of 195 documents, with 11,2k annotated candidate relations, the Dev section of 98 documents and 6,2 annotated candidates, and the Test section contains 100 documents and 5,9k candidates. We now refer to this dataset as $D^{rc}$.

As data for the SG(LR) loss, we use the raw THYME train texts, extended with a MIMIC III \cite{johnson2016mimic} section of 500 discharge summaries that contain the terms 'colon' and 'cancer' at least twice. We refer to this dataset as $D^{sg}$ and it is used in all pre-training and joint training settings.

\subsection{Training Settings}
We compare five training settings for the model of which the first three settings act as baselines to compare with our proposed models (settings 4 and 5):
\begin{enumerate}
\item \textbf{RC (random initialization):} Uses random word embedding initializations (picked from [0.05, 0.05]) and trains on loss $L_{rc}$.
\item \textbf{RC (SG initialization):} Initializes the model with pre-trained SG embeddings, and trains on $L_{rc}$.
\item \textbf{RC (SG fixed):} Initializes the model with pre-trained SG embeddings, and trains the model on $L_{rc}$, while not updating the word embedding weights, keeping them as fixed features.
\item \textbf{RC + SG:} Initializes the model with pre-trained SG embeddings, and trains on $L_{rc+sg}$.
\item \textbf{RC + SGLR:} Initializes the model with pre-trained SG embeddings, and trains on $L_{rc+sglr}$.
\end{enumerate}

\subsection{Evaluation}
As evaluation metrics we use precision, recall, and F-measure, calculated using the evaluation script provided by the Clinical TempEval organizers\footnote{https://github.com/bethard/anaforatools}, which evaluates the CR under the temporal closure \cite{uzzaman2012tempeval}, taking into account transitivity properties of the temporal relations.

\subsection{Preprocessing \& Hyper-parameters}
\label{sec:preproc}
As preprocessing of the corpus, we employ very simple tokenization: splitting the text on spaces and considering punctuation\footnote{\texttt{$,./\backslash$"'=+-;:()!?<>\%\&\$*|[]\{\}}} and newlines as individual tokens. Additionally, we lowercase the corpus, and conflate digits ($1992 \rightarrow 5555$). To extract POS we use the Stanford POS Tagger v3.7 \cite{toutanova2003feature}, using the pre-trained (on WSJ) caseless left-3-words model. Finally, all 1-time occurring tokens in the training dataset are replaced by a \mbox{\textless \sc{unk}\textgreater}-token, to represent out-of-vocabulary words at test-time.

We employ the same candidate generation as \citet{leeuwenberg2017structured}, considering all pairs of events (Event$\times$Event, or EE) and events and temporal expressions (Timex3$\times $Event, or TE) with a maximum token distance of 30 as candidate relations to be classified (ignoring sentence boundaries, as relations also occur across them). This candidate generation has a maximum recall of \blue{0.87\%}, and gives a ratio between the positive and negative class of 1:36, also indicating the task's difficulty.

In our experiments, we tuned each model type within the same hyper-parameter search space on the Dev set. The number of LSTM units was chosen from $\{25, 50, 100\}$, and the word embedding dimension from $\{25, 50, 100\}$. This resulted in \blue{100} LSTM units, and a word embedding size of \blue{25}. The loss weights $\lambda_{sg}$ and $\lambda_{sglr}$
were chosen from $\{0.01, 0.1, 1.0, 10, 100\}$, resulting in 
\blue{$\lambda_{sg}$=0.1 and $\lambda_{sglr}$=0.1}. The context window size of the skip-gram was set to \blue{2}, chosen from $\{2, 4, 8\}$. The context size for the RC, and POS embedding dimension size were not tuned and set to 10 (left and right), and \blue{40} respectively.

\section{Results}
\definecolor{ao}{rgb}{0.0, 0.5, 0.0}

\subsection{Influence of Word-Level Loss}
We looked at model performance when increasing the importance of the auxiliary word-level loss ($\lambda_{sg}$ and $\lambda_{sglr}$). The results when changing these hyper-parameters are shown in Fig. \ref{fig:lambda}.
\begin{figure}[!ht]
    \centering
    \begin{minipage}{.48\textwidth}
        \centering	
\resizebox{.9\textwidth}{!}{        
\begin{tikzpicture}
    \begin{axis}
        [
        xlabel={$\lambda$},
        ylabel={},
        xtick = {0,1,2,3,4,5},
  		xticklabels = {0,0.01,0.1,1.0, 10, 100},
        grid=both,
        grid style={line width=.1pt, draw=gray!10}
        ]
        
    \draw (0,140) node [color=Blue] {P};
    \addplot[color=blue,mark=*,smooth, dashed] coordinates
    {(0,62) (1,65.9) (2,67.1) (3,66.4) (4,67.4) (5,68.7)};
    \addplot[color=blue,mark=diamond*,smooth, dashed] coordinates
    {(0,62) (1,63.6) (2,63.2) (3,69.7) (4,69.9) (5,67.8)};
    
    \draw (0,45) node [color=red] {R};    
    \addplot[color=red,mark=*,smooth, dashed] coordinates
    {(0,53.3) (1,55.5) (2,55.3) (3,55.3) (4,53) (5,51.4)};
    \addplot[color=red,mark=diamond*,smooth, dashed] coordinates
    {(0,53.3) (1,55.6) (2,59.5) (3,52.5) (4,50.7) (5,52.1)};

    \draw (0,90) node [color=Green] {F};
    \addplot[color=Green,mark=*,smooth] coordinates
    {(0,57.3) (1,60.3) (2,60.6) (3,60.3) (4,59.4) (5,58.8)};
    \addplot[color=Green,mark=diamond*,smooth] coordinates
    {(0,57.3) (1,59.3) (2,61.3) (3,59.9) (4,58.7) (5,58.9)};

    \end{axis}
\end{tikzpicture}}
\caption{\label{fig:lambda} Precision (P), Recall (R) and F-measure (F) on the THYME Dev set for different values of $\lambda_{sg}$ ($\circ$) and $\lambda_{sglr}$ ($\diamond$).}
    \end{minipage}%
    \hfill
    \begin{minipage}{.48\textwidth}
        \centering
\resizebox{\textwidth}{!}{
\begin{tikzpicture}
    \begin{axis}
        [
        xlabel={Training Set Size},
        ylabel shift = -.05cm,
        ylabel={F-measure},
        xtick = {0,1,2,3,4},
  		xticklabels = {20$\%$,40$\%$,60$\%$,80$\%$,100$\%$},
        legend pos=south east,
        legend style={font=\scriptsize},
        legend cell align={left},
        grid=both,
        grid style={line width=.1pt, draw=gray!10}
        ]
        
    \addplot[color=Green,mark=*,smooth] coordinates
    {(0,54) (1,57) (2,58.7) (3,61) (4,60.2)};
	\addlegendentry{RC + SG};

    \addplot[color=blue,mark=diamond*,smooth] coordinates
    {(0,54.6) (1,56.9) (2,60.1) (3,59.5) (4,60.2)};
	\addlegendentry{RC + SGLR};
   
    \addplot[color=purple,mark=square*,smooth] coordinates
    {(0,51.9) (1,55.1) (2,57.1) (3,58.7) (4,59.2)};  
	\addlegendentry{RC (SG fixed)};

    \addplot[color=orange,smooth,mark=triangle*] coordinates
    {(0,52.1) (1,54.8) (2,56.6) (3,57.9) (4,58.9)};
	\addlegendentry{RC (SG init.)};

    \addplot[color=red,smooth, mark=*] coordinates
    {(0,48.5) (1,50.4) (2,53.5) (3,53.6) (4,55.6)};
	\addlegendentry{RC (rnd. init.)};

\end{axis}
    
\end{tikzpicture}}
\caption{\label{fig:trainsetsize} F-measure on the THYME Dev set for different training settings, over different training set sizes (in \% of the full train set).}
    \end{minipage}
\end{figure}

When choosing $\lambda=0$ ($\lambda_{sg}$ or $\lambda_{sglr}$) for our model, we obtain the same model as RC (SG init.), as the auxiliary objective has no influence.
For very high values of $\lambda$, we hypothesize that the models converge towards RC (SG fixed), because when taking $\lambda \rightarrow \infty$, the word embeddings are solely optimized for the auxiliary loss, as the influence of the task loss is proportionally zero, i.e. lim$_{\lambda \rightarrow \infty} \frac{1}{\lambda} = 0$.
This property is interesting, as it shows these baseline models can act as lower bounds for our model performance when choosing a bad $\lambda$ value.
This can be observed in Fig. \ref{fig:lambda}, where the F-measure is highest for a $\lambda$ that balance both objectives, whereas for extremes F-measure decreases.

\subsection{Comparison Across Training Set Size}
We evaluated all model settings for different training set sizes.
From Fig.\ref{fig:trainsetsize} we can see that the worst model is the one with random word embedding initializations. One improvement is to initialize the model with pre-trained SG embeddings. Fixing the pre-trained embeddings or continued training on the main task objective does not result in very different F-measure scores. However, continued training on the combined objective does seem to give a significant increase in F-measure, consistent over different training sizes for both the RC + SG as well as the RC + SGLR variant. Additionally, it should be noticed that parameters are not returned on each dataset size, but \addedd{obtained} from tuning on the full Dev set. Still the model ranking is consistent.

\subsection{Evaluation on Subsets of Relations}
\label{sec:subsets}
To get a more detailed insight in what each model learns relative to the others, we evaluated our models on different subsets of the data. \added{First, we split the containment relations based on their argument types} and separately evaluated the 3.3k EE relations and the 2.7k TE relations. EE relations are generally found more difficult than TE relations \cite{lin2015multilayered,lin2016improving,dligach2017neural,lin2017representations}. In Table \ref{tab:filtered_evaluation} we can see that also for our model, EE relations \addedd{are harder to recognize} than the TE relations, as all models achieve higher scores for TE compared to EE relations. 
\begin{table}[h!]
\centering
\caption{\label{tab:filtered_evaluation} Evaluation on subsets of THYME Dev (in F-measure). The subsets of Event$\times$Event (EE) and Timex3$\times$Event (TE) relation pairs are of sizes 3.3k and 2.7k respectively. The intervals 0-100, 100-500 and 500+ are subsets reflecting average argument token frequency in the training data (of sizes 2.2k, 2.2k and 1.8k respectively).}
\begin{tabular}{lcccccccc}
\topline
Model &  EE & TE & 0-100 & 100-500 & 500+ & All \\ 
\midline
RC (random initializations) & 44.5 & 64.4 & 40.5 & 57.8 & 63.4 & 53.4 \\
RC (SG initializations) & 49.5 & 68.6 & 44.1 & 62.5 & 67.0 & 57.3 \\
RC (SG fixed) & 48.9 & \textbf{68.7} & 44.1 & 62.7 & 67.3 & 57.6 \\ 
\midline
RC + SG & 51.6 & 67.4 & \textbf{46.4} & 62.5 & 66.8 & 58.2 \\
RC + SGLR & \textbf{51.7} & 68.5 & 45.3 & \textbf{63.0} & \textbf{68.1} & \textbf{58.4} \\
\bottomline
\end{tabular}
\vspace{-.3cm}
\end{table}
What is interesting to see is that when training with the combined loss (SG or SGLR) we obtain a clear improvement on the more difficult EE relations, and perform slightly worse on TE relations compared to using pre-trained embeddings (the three upper settings). The reason could be that EE relations are more diverse in vocabulary, and are consequently more influenced by the quality of the embeddings.

We also analyzed the models w.r.t. total frequency in the training data ($D^{rc} + D^{sg}$) and made three subsets based on the average word frequency of the argument tokens in each relation. The three buckets of relations, 0-100, 100-500, and 500+, are of sizes 2.2k, 2.2k, and 1.8k respectively. What can be observed is that the RC+SG model performs best for low-frequency words, and RC+SGLR performs best for the higher frequency ranges. This can be explained by the fact that the SGLR separates left and right context words, creating sparser and more precise contexts compared to SG. Sparse context descriptions can hurt representations of low frequency words as there may not be enough words that share contexts. But, for more frequent words, more precise context descriptions as in SGLR help to prevents incorrect generalizations (such as cases where word order matters). 
When evaluating on the full Dev set, both combined loss settings outperform the baselines consistently.


\subsection{Comparison to the State of the Art}
We also compared our proposed models to various state-of-the-art systems from the literature:

The THYME system, by \citet{lin2016improving}, consist of separate models for EE relations and TE relations. They employ two feature rich support vector machines (SVM), using POS and dependency parse features from the cTAKES clinical pipeline \cite{savova2010mayo} together with augmented training through extended UMLS entities. They later replaced the TE component by a token-based CNN model which improved their model \cite{lin2017representations,dligach2017neural}. Also replacing the EE component by a CNN model decreased model performance, showing that the CNN was not able to replace the feature rich SVM.
\citet{leeuwenberg2017structured} used a feature rich structured perceptron, also using cTAKES POS and dependency parse features, jointly learning different relation types on the document level.
\citet{tourille2017bilstm} used two bidirectional LSTM models, one for inter-sentence and one for intra-sentence relations. They used fixed word embeddings pre-trained on the MIMIC III corpus, and also incorporated character level information. To obtain their top results they added ground truth event attribute features enhanced with entity information also obtained from cTAKES.

As can be noticed, all state-of-the-art baselines used dedicated clinical NLP tools to enhance their features in order to obtain their top results, in contrast to our model, which uses only the Stanford POS Tagger (trained on news texts).
\begin{table}[h!]
\centering
\caption{\label{tab:results_test}THYME test set results, reporting precision (P), recall (R) and F-measure (F), macro-averaged over three runs. The standard deviation for F is also given.}

\begin{tabular}{lccc}
\topline
Model & P & R & F \\ \midline
\textit{With specialized resources:}\\
Best Clinical TempEval (2016) & 58.8 & 55.9 &  57.3\\
Lin et al. (2016) & 66.9 & 53.4 & 59.4 \\
Leeuwenberg et al. (2017) & - & - & 60.8 \\ 
Tourille et al. (2017) & 65.7 & 57.5 & 61.3 \\ 
Lin et al. (2017) & 66.2 & 58.5 & 62.1\\ 
\midrule
\textit{No specialized resources:}\\
RC (random initialization) & 67.9 & 52.1 & 58.9\tiny$\pm0.2$\\
RC (SG initialization)& \textbf{71.2} & 52.0 & 60.0\tiny$\pm1.2$\\
RC (SG fixed)& 68.9 & 54.6 & 60.9\tiny$\pm0.8$\\
RC + SG & 66.2 & \textbf{59.7} & \textbf{62.8}\tiny$\pm0.2$\\	
RC + SGLR & 68.7 & 57.5 & 62.5\tiny$\pm0.3$\\
\bottomline
\end{tabular}
\end{table}

Table \ref{tab:results_test} shows that initializing the model with the pre-trained embeddings gives a significant \footnote{$P<0.0001$ for a document-level pairwise t-test} \blue{1.1} point increase in F-measure compared to random initialization, due to an increase in precision. Fixing the embeddings gives slightly better performance than using them as initialization, an increase of \blue{0.9} point in F-measure, mostly due to higher recall. When extending the loss with the SGLR loss, we gain\footnotemark[6] \blue{1.6} in F-measure compared to fixing the word embeddings, and also surpass the state of the art by \blue{0.4} even without specialized resources. If we train our model using the SG loss extension we obtain the best results, and gain\footnotemark[6] \blue{1.9} points in F-measure compared to using pre-trained fixed word embeddings. This setting also exceeds the state of the art \cite{lin2017representations} by \blue{0.7} points in F-measure, due to a gain of \blue{1.2} points in recall, again without using any specialized clinical NLP tools for feature engineering, in contrast to all state-of-the-art baselines.
\subsection{Manual Error Analysis}
Finally, we manually analyzed 50 false positives and 50 false negatives picked randomly from the test set predictions for different settings.

From Table \ref{tab:error_analysis} we can see that all models have difficulties with distant relations that cross sentence or clause boundaries (CCR). This could be because class imbalance correlates with distance between the arguments of the temporal relations. Furthermore, arguments that are frequent in the supervised data ($>250$) are a dominant error category. We suspect this is because frequent events often function both as container and as contained, whereas infrequent events are less ambiguous in their argument position. This hurts RC (SG fixed) most as its embeddings are not influenced by $D^{rc}$.
Furthermore it can be noticed that RC+SG has less errors for infrequent arguments ($< 10$) in the supervised data. This could be because it leverages the few available instances from both the $D^{rc}$ and $D^{sg}$ data better than the single-loss models.
\begin{table}[h!]
\caption{\label{tab:error_analysis}Error analysis on 50 FP and 50 FN (random from test) for different settings. Clause boundaries are: newlines and sub-clause or sentence boundaries. Error categories are not mutually exclusive. 
}
\centering
\begin{tabular}{lccc}
\topline
Error Type & RC + SG & RC (SG fixed) & RC (SG init.) \\
 \midline
Cross-Clause Relations (CCR) & 42 & 39 & 36 \\
Infrequent Arguments ($<10$)& 11 & 15 & 26 \\
Frequent Arguments ($>250$) & 37 & 50 & 40 \\
Mistake in Ground-Truth & 10 & 8 & 5 \\
Other & 21 & 15 & 28 \\
\bottomline
\end{tabular}
\end{table}
\section{Conclusions}
In this work, we proposed a neural relation classification model for the extraction of narrative containment relations from clinical texts.\footnote{Code is available at: \url{http://liir.cs.kuleuven.be/software.html}}

The model trains word representations jointly on the \addet{supervised} \addet{relation classification} task and an \addet{unsupervised} auxiliary skip-gram objective (with separate datasets) through weight sharing to more effectively exploit both the unlabeled and labeled data, as annotated clinical data is costly to create. We show that this word-level joint training results in significantly better generalizing classification models compared to using pre-trained word embeddings (either as initialization or fixed embeddings).
Furthermore, we show that performance trends and good values for $\lambda$ (balance between tasks) are robust over different training set sizes, and that even for (badly tuned) extreme values of $\lambda$ the quality of the model's embeddings is naturally lower-bounded by their pre-trained variants. Additionally, our model sets a new state of the art for temporal relation extraction on the THYME dataset, without using extra dedicated clinical resources, in contrast to current state-of-the-art models.

As future work, it would be interesting to see how well the improvements caused by the word-level joint training generalize to other NLP tasks that typically use pre-trained word embeddings.

\section*{Acknowledgment}
The authors would like to thank the reviewers for their constructive comments which helped us to improve the paper. Also, we would like to thank the Mayo Clinic for permission to use the THYME corpus. This work was funded by the KU Leuven C22/15/16 project "MAchine Reading of patient recordS (MARS)", and by the IWT-SBO 150056 project "ACquiring CrUcial Medical information Using LAnguage TEchnology" (ACCUMULATE).




\nocite{lee2016uthealth}
\bibliography{references}

\begin{thebibliography}{}

\bibitem[\protect\citename{Baxter and others}2000]{baxter2000model}
Jonathan Baxter et~al.
\newblock 2000.
\newblock A model of inductive bias learning.
\newblock {\em Journal of Artificial Intelligence Research}, 12(149-198):3.

\bibitem[\protect\citename{Bethard \bgroup et al.\egroup
  }2015]{bethard2015semeval}
Steven Bethard, Leon Derczynski, Guergana Savova, James Pustejovsky, and Marc
  Verhagen.
\newblock 2015.
\newblock {SemEval}-2015 task 6: {Clinical TempEval}.
\newblock In {\em Proc. of SemEval}, pages 806--814. ACL.

\bibitem[\protect\citename{Bethard \bgroup et al.\egroup
  }2016]{bethard2016semeval}
Steven Bethard, Guergana Savova, Wei-Te Chen, Leon Derczynski, James
  Pustejovsky, and Marc Verhagen.
\newblock 2016.
\newblock {SemEval}-2016 task 12: {Clinical} {TempEval}.
\newblock {\em Proc. of SemEval}, pages 1052--1062.

\bibitem[\protect\citename{Bethard \bgroup et al.\egroup
  }2017]{bethard17semeval}
Steven Bethard, Guergana Savova, Martha Palmer, and James Pustejovsky.
\newblock 2017.
\newblock {SemEval}-2017 task 12: {Clinical TempEval}.
\newblock In {\em Proc. of SemEval}, pages 565--572, Vancouver, Canada, August.
  ACL.

\bibitem[\protect\citename{Caruana}1998]{caruana1998multitask}
Rich Caruana.
\newblock 1998.
\newblock Multitask learning.
\newblock In {\em Learning to Learn}, pages 95--133. Springer.

\bibitem[\protect\citename{Cheng \bgroup et al.\egroup }2015]{cheng2015open}
Hao Cheng, Hao Fang, and Mari Ostendorf.
\newblock 2015.
\newblock Open-domain name error detection using a multi-task {RNN}.
\newblock In {\em Proc. of EMNLP}, pages 737--746.

\bibitem[\protect\citename{Collobert and Weston}2008]{collobert2008unified}
Ronan Collobert and Jason Weston.
\newblock 2008.
\newblock A unified architecture for natural language processing: Deep neural
  networks with multitask learning.
\newblock In {\em Proceedings of the 25th international conference on Machine
  learning}, pages 160--167. ACM.

\bibitem[\protect\citename{Derczynski}2017]{derczynski2016automatically}
Leon~RA Derczynski.
\newblock 2017.
\newblock Automatically ordering events and times in text.
\newblock In {\em Studies in Computational Intelligence}, volume 677. Springer.

\bibitem[\protect\citename{Dligach \bgroup et al.\egroup
  }2017]{dligach2017neural}
Dmitriy Dligach, Timothy Miller, Chen Lin, Steven Bethard, and Guergana Savova.
\newblock 2017.
\newblock Neural temporal relation extraction.
\newblock {\em Proc. of EACL}, page 746.

\bibitem[\protect\citename{Dong \bgroup et al.\egroup }2015]{dong2015multi}
Daxiang Dong, Hua Wu, Wei He, Dianhai Yu, and Haifeng Wang.
\newblock 2015.
\newblock Multi-task learning for multiple language translation.
\newblock In {\em Proc. of ACL}, pages 1723--1732. ACL.

\bibitem[\protect\citename{Hashimoto \bgroup et al.\egroup
  }2017]{hashimoto2016joint}
Kazuma Hashimoto, Caiming Xiong, Yoshimasa Tsuruoka, and Richard Socher.
\newblock 2017.
\newblock A joint many-task model: Growing a neural network for multiple {NLP}
  tasks.
\newblock {\em Proc. of EMNLP}.

\bibitem[\protect\citename{Hochreiter and Schmidhuber}1997]{hochreiter1997long}
Sepp Hochreiter and J{\"u}rgen Schmidhuber.
\newblock 1997.
\newblock Long short-term memory.
\newblock {\em Neural Computation}, 9(8):1735--1780.

\bibitem[\protect\citename{Jiang \bgroup et al.\egroup }2015]{jiang2015parsing}
Min Jiang, Yang Huang, Jung-wei Fan, Buzhou Tang, Josh Denny, and Hua Xu.
\newblock 2015.
\newblock Parsing clinical text: how good are the state-of-the-art parsers?
\newblock {\em BMC Medical Informatics and Decision Making}, 15(1):S2.

\bibitem[\protect\citename{Johnson \bgroup et al.\egroup
  }2016]{johnson2016mimic}
Alistair~EW Johnson, Tom~J Pollard, Lu~Shen, Li-wei~H Lehman, Mengling Feng,
  Mohammad Ghassemi, Benjamin Moody, Peter Szolovits, Leo~Anthony Celi, and
  Roger~G Mark.
\newblock 2016.
\newblock {MIMIC-III}, a freely accessible critical care database.
\newblock {\em Scientific Data}, 3.

\bibitem[\protect\citename{Kingma and Ba}2014]{kingma2014adam}
Diederik~P. Kingma and Jimmy Ba.
\newblock 2014.
\newblock Adam: A method for stochastic optimization.
\newblock In {\em Proc. of ICLR}.

\bibitem[\protect\citename{Lee \bgroup et al.\egroup }2016]{lee2016uthealth}
Hee-Jin Lee, Hua Xu, Jingqi Wang, Yaoyun Zhang, Sungrim Moon, Jun Xu, and
  Yonghui Wu.
\newblock 2016.
\newblock Uthealth at {SemEval}-2016 task 12: an end-to-end system for temporal
  information extraction from clinical notes.
\newblock In {\em Proc. of SemEval}, pages 1292--1297. ACL.

\bibitem[\protect\citename{Leeuwenberg and
  Moens}2017]{leeuwenberg2017structured}
Artuur Leeuwenberg and Marie-Francine Moens.
\newblock 2017.
\newblock Structured learning for temporal relation extraction from clinical
  records.
\newblock In {\em Proc. of EACL}, pages 1150--1158. ACL.

\bibitem[\protect\citename{Lin \bgroup et al.\egroup
  }2015]{lin2015multilayered}
Chen Lin, Dmitriy Dligach, Timothy~A Miller, Steven Bethard, and Guergana~K
  Savova.
\newblock 2015.
\newblock Multilayered temporal modeling for the clinical domain.
\newblock {\em Journal of the American Medical Informatics Association},
  23(2):387--395.

\bibitem[\protect\citename{Lin \bgroup et al.\egroup }2016]{lin2016improving}
Chen Lin, Timothy Miller, Dmitriy Dligach, Steven Bethard, and Guergana Savova.
\newblock 2016.
\newblock Improving temporal relation extraction with training instance
  augmentation.
\newblock In {\em Proc. of ACL}, page 108. ACL.

\bibitem[\protect\citename{Lin \bgroup et al.\egroup
  }2017]{lin2017representations}
Chen Lin, Timothy Miller, Dmitriy Dligach, Steven Bethard, and Guergana Savova.
\newblock 2017.
\newblock Representations of time expressions for temporal relation extraction
  with convolutional neural networks.
\newblock In {\em Proc. of BioNLP}, page 322. ACL.

\bibitem[\protect\citename{Ling \bgroup et al.\egroup }2015]{ling2015two}
Wang Ling, Chris Dyer, Alan~W Black, and Isabel Trancoso.
\newblock 2015.
\newblock Two/too simple adaptations of word2vec for syntax problems.
\newblock In {\em Proc. of HLT-NAACL}, pages 1299--1304. ACL.

\bibitem[\protect\citename{Mikolov \bgroup et al.\egroup
  }2013]{mikolov2013efficient}
Tomas Mikolov, Kai Chen, Greg Corrado, and Jeffrey Dean.
\newblock 2013.
\newblock Efficient estimation of word representations in vector space.
\newblock {\em arXiv preprint arXiv:1301.3781}.

\bibitem[\protect\citename{Morgan and Bourlard}1990]{morgan1990generalization}
Nelson Morgan and Herv{\'e} Bourlard.
\newblock 1990.
\newblock Generalization and parameter estimation in feedforward nets: Some
  experiments.
\newblock In {\em Advances in Neural Information Processing Systems}.

\bibitem[\protect\citename{Nguyen and Grishman}2015]{nguyen2015relation}
Thien~Huu Nguyen and Ralph Grishman.
\newblock 2015.
\newblock Relation extraction: Perspective from convolutional neural networks.
\newblock In {\em Proc. of NAACL-HLT}, pages 39--48.

\bibitem[\protect\citename{Onisko \bgroup et al.\egroup }2015]{Onisko2015}
Agnieszka Onisko, Allan Tucker, and Marek~J. Druzdzel.
\newblock 2015.
\newblock Prediction and prognosis of health and disease.
\newblock In {\em Foundations of Biomedical Knowledge Representation: Methods
  and Applications}, pages 181--188. Springer.

\bibitem[\protect\citename{Peng and Dredze}2015]{peng2015named}
Nanyun Peng and Mark Dredze.
\newblock 2015.
\newblock Named entity recognition for chinese social media with jointly
  trained embeddings.
\newblock In {\em Proceedings of the 2015 Conference on Empirical Methods in
  Natural Language Processing}, pages 548--554.

\bibitem[\protect\citename{Petrov \bgroup et al.\egroup }2012]{tagsetlrec2012}
Slav Petrov, Dipanjan Das, and Ryan McDonald.
\newblock 2012.
\newblock A universal part-of-speech tagset.
\newblock In {\em Proc. of LREC}, pages 2089--2096. European Language Resources
  Association (ELRA), May.

\bibitem[\protect\citename{Ruder}2017]{ruder2017overview}
Sebastian Ruder.
\newblock 2017.
\newblock An overview of multi-task learning in deep neural networks.
\newblock {\em arXiv preprint arXiv:1706.05098}.

\bibitem[\protect\citename{Savova \bgroup et al.\egroup }2010]{savova2010mayo}
Guergana~K Savova, James~J Masanz, Philip~V Ogren, Jiaping Zheng, Sunghwan
  Sohn, Karin~C Kipper-Schuler, and Christopher~G Chute.
\newblock 2010.
\newblock Mayo clinical text analysis and knowledge extraction system
  ({cTAKES}): architecture, component evaluation and applications.
\newblock {\em Journal of the American Medical Informatics Association},
  17(5):507--513.

\bibitem[\protect\citename{S{\o}gaard and Goldberg}2016]{sogaard2016deep}
Anders S{\o}gaard and Yoav Goldberg.
\newblock 2016.
\newblock Deep multi-task learning with low level tasks supervised at lower
  layers.
\newblock In {\em Proc. of ACL}, volume~2, pages 231--235. ACL.

\bibitem[\protect\citename{Srivastava \bgroup et al.\egroup
  }2014]{srivastava2014dropout}
Nitish Srivastava, Geoffrey~E Hinton, Alex Krizhevsky, Ilya Sutskever, and
  Ruslan Salakhutdinov.
\newblock 2014.
\newblock Dropout: a simple way to prevent neural networks from overfitting.
\newblock {\em Journal of Machine Learning Research}, 15(1):1929--1958.

\bibitem[\protect\citename{Styler~IV \bgroup et al.\egroup
  }2014]{styler2014temporal}
William~F Styler~IV, Steven Bethard, Sean Finan, Martha Palmer, Sameer Pradhan,
  Piet~C de~Groen, Brad Erickson, Timothy Miller, Chen Lin, Guergana Savova,
  et~al.
\newblock 2014.
\newblock Temporal annotation in the clinical domain.
\newblock {\em Transactions of the Association for Computational Linguistics},
  2:143--154.

\bibitem[\protect\citename{Sun \bgroup et al.\egroup }2013]{sun2013evaluating}
Weiyi Sun, Anna Rumshisky, and Ozlem Uzuner.
\newblock 2013.
\newblock Evaluating temporal relations in clinical text: 2012 i2b2 challenge.
\newblock {\em Journal of the American Medical Informatics Association},
  20(5):806--813.

\bibitem[\protect\citename{Tourille \bgroup et al.\egroup
  }2017]{tourille2017bilstm}
Julien Tourille, Olivier Ferret, Aurelie Neveol, and Xavier Tannier.
\newblock 2017.
\newblock Neural architecture for temporal relation extraction: A {Bi-LSTM}
  approach for detecting narrative containers.
\newblock In {\em Proc. of ACL}, pages 224--230, Vancouver, Canada, July. ACL,
  ACL.

\bibitem[\protect\citename{Toutanova \bgroup et al.\egroup
  }2003]{toutanova2003feature}
Kristina Toutanova, Dan Klein, Christopher~D Manning, and Yoram Singer.
\newblock 2003.
\newblock Feature-rich part-of-speech tagging with a cyclic dependency network.
\newblock In {\em Proc. of NAACL-HLT}, pages 173--180. ACL.

\bibitem[\protect\citename{UzZaman \bgroup et al.\egroup
  }2012]{uzzaman2012tempeval}
Naushad UzZaman, Hector Llorens, James Allen, Leon Derczynski, Marc Verhagen,
  and James Pustejovsky.
\newblock 2012.
\newblock Tempeval-3: Evaluating events, time expressions, and temporal
  relations.
\newblock {\em arXiv preprint arXiv:1206.5333}.

\bibitem[\protect\citename{Yu and Jiang}2016]{yu2016learning}
Jianfei Yu and Jing Jiang.
\newblock 2016.
\newblock Learning sentence embeddings with auxiliary tasks for cross-domain
  sentiment classification.
\newblock In {\em Proc. of EMNLP}, pages 236--246. ACL, November.

\bibitem[\protect\citename{Zeng \bgroup et al.\egroup }2014]{zeng2014relation}
Daojian Zeng, Kang Liu, Siwei Lai, Guangyou Zhou, Jun Zhao, et~al.
\newblock 2014.
\newblock Relation classification via convolutional deep neural network.
\newblock In {\em Proc. of COLING}, pages 2335--2344. ACL.

\bibitem[\protect\citename{Zhang and Wang}2015]{zhang2015relation}
Dongxu Zhang and Dong Wang.
\newblock 2015.
\newblock Relation classification via recurrent neural network.
\newblock {\em arXiv preprint arXiv:1508.01006}.

\bibitem[\protect\citename{Zhang \bgroup et al.\egroup
  }2015]{zhang2015bidirectional}
Shu Zhang, Dequan Zheng, Xinchen Hu, and Ming Yang.
\newblock 2015.
\newblock Bidirectional long short-term memory networks for relation
  classification.
\newblock In {\em Proc. of PACLIC}.

\bibitem[\protect\citename{Zhou \bgroup et al.\egroup }2016]{zhou2016attention}
Peng Zhou, Wei Shi, Jun Tian, Zhenyu Qi, Bingchen Li, Hongwei Hao, and Bo~Xu.
\newblock 2016.
\newblock Attention-based bidirectional long short-term memory networks for
  relation classification.
\newblock In {\em Proc. of ACL}.

\end{thebibliography}
\bibliographystyle{acl}

\end{document}